# Modeling and Inverse Identification of Interfacial Heat Conduction in Finite Layer and Semi-Infinite Substrate Systems via a Physics-Guided Neural Framework

[1]Wenhao Sha, [1,2,3]*Tienchong Chang

[1]Shanghai Institute of Applied Mathematics and Mechanics, Shanghai Key Laboratory of Mechanics in Energy Engineering, Shanghai Frontier Science Center of Mechanoinformatics, School of Mechanics and Engineering Science, Shanghai University, Shanghai 200072, China

[2]Shanghai Institute of Aircraft Mechanics and Control, Tongji University, Shanghai 200092, China

[3]Joint-Research Center for Computational Materials, Zhejiang Laboratory, Hangzhou 311100, China

**Abstract**

Heat transfer in semiconductor devices is dominated by chip and substrate assemblies, where heat generated within a finite chip layer dissipates into a semi-infinite substrate with much higher thermophysical properties. This mismatch produces steep interfacial temperature gradients, making the transient thermal response highly sensitive to the interface. Conventional numerical solvers require excessive discretization to resolve these dynamics, while physics-informed neural networks (PINNs) often exhibit unstable convergence and loss of physical consistency near the material interface. To address these challenges, we introduce HeatTransFormer, a physics-guided Transformer architecture for interface-dominated diffusion problems. The framework integrates physically informed spatiotemporal sampling, a Laplace-based activation emulating analytical diffusion solutions, and a mask-free attention mechanism supporting bidirectional spatiotemporal coupling. These components enable the model to resolve steep gradients, maintain physical consistency, and remain stable where PINNs typically fail. HeatTransFormer produces coherent temperature fields across the interface when applied to a finite layer and semi-infinite substrate configuration. Coupled with a physics-constrained inverse strategy, it further enables reliable identification of three unknown thermal properties simultaneously using only external measurements. Overall, this work demonstrates that physics-guided Transformer architectures provide a unified framework for forward and inverse modeling in interface-dominated thermal systems.

* To whom correspondence should be addressed, E-mail: *tchang@staff.shu.edu.cn*

## 1. Introduction

The rapid scaling of semiconductor devices has led to densely stacked multilayer structures where heat must traverse several thermally mismatched material interfaces [1-3]. These interfaces govern the formation of hotspots, the dissipation of heat, and ultimately the reliability of modern chips. As power densities grow, understanding and accurately modeling heat transfer through such multilayer assemblies has become increasingly critical for next-generation electronic systems [4-7].

In many packaged devices, as illustrated in **Fig. 1(a)**, heat generated within the chip flows predominantly in the vertical direction into the heat-spreading substrate or heat sink. Owing to the large disparity between the lateral dimensions and the chip thickness, together with the high thermal conductivity of the underlying substrate, the system can be rigorously reduced to the one-dimensional configuration [8-9] shown in **Fig. 1(b)**, where the chip behaves as a finite layer and the substrate functions as a semi-infinite medium. This classical layer and substrate abstraction captures the essential interfacial thermal physics while enabling efficient analysis of transient conduction.

Within this reduced representation, the effective thermal conductivity and diffusivity of the chip layer become the primary determinants of junction temperature, hotspot severity, and allowable power dissipation [10-11]. Therefore, these effective thermophysical properties are indispensable for thermal model calibration, system-level simulation, and reliability prediction [2]. However, despite their importance, these properties cannot be measured directly once the device is packaged, because the chip becomes enclosed beneath multiple structural layers that block sensing access, while destructive testing is unacceptable for modern devices due to their high value and stringent reliability requirements [12-14].

Therefore, obtaining the chip's effective thermophysical properties requires non-destructive inference from external temperature measurements, and this capability has become increasingly important for advanced semiconductor qualification and thermal design workflows. Consequently, the task of determining the chip's effective thermal properties becomes an inverse heat conduction problem in which the chip's internal thermal behavior must be reconstructed from substrate-side transient measurements [4, 6].

However, solving this inverse problem is fundamentally difficult. The abrupt change in thermal properties at the interface forces the temperature field to develop steep gradients [8,15], and these gradients dominate the transient response in the early stages of heat diffusion. At the same time, thermal conductivity and diffusivity shape the temperature evolution through mechanisms that are closely intertwined, which makes it difficult to isolate their individual contributions from a single surface measurement.

Because of these combined effects, even small perturbations in the measured data can lead to significant errors, and the estimation process becomes highly ill-posed [4, 16].

Traditional numerical approaches such as the finite element method (FEM) [17] or finite difference method (FDM) [18] can accurately compute forward heat conduction but become computationally prohibitive for inverse characterization due to the need for repeatedly solving partial differential equations (PDEs) and extremely fine interface discretization [19-20]. Physics-informed neural networks (PINNs) [21-24] alleviate data requirements by embedding governing equations into the loss function, yet standard feedforward architectures struggle to represent the steep interfacial gradients [25-26] and coupled transient dynamics seen in layer and substrate systems [27-28], often resulting in unstable [29] or non-convergent inverse predictions [30-31]. These limitations indicate that neither conventional solvers nor PINNs provide the inductive biases necessary for reliable interface-dominated thermal inversion. In contrast, Transformer architectures [32] exhibit strong capabilities for processing structured sequences, and this makes them naturally suitable for capturing the steep temperature gradients that arise at material interfaces, which can be viewed as short, localized spatiotemporal patterns requiring precise and context-aware propagation. However, the way Transformers construct sequences [33-35], choose activations, and apply masks is still tailored to text and vision tasks rather than heat conduction, and these design choices break down when applied to the strong gradients and coefficient jumps at material interfaces.

To address this gap, we develop HeatTransFormer (HTF), a decoder-only Transformer architecture for transient heat conduction modeling, particularly in systems where interfacial thermal discontinuities dominate the dynamics. HTF incorporates three physics-motivated components. First, an anchor-based spatiotemporal sampling strategy constructs local sequences centered at critical spatial anchors, especially the material interface, which allows the model to capture sharp temperature-gradient transitions. Second, we introduce LaplaceAct, an activation function inspired by the inverse Laplace transform, which embeds the exponential-decay and oscillatory modes characteristic of analytical heat diffusion solutions, thereby enhancing both physical interpretability and representational efficiency. Third, a mask-free self-attention mechanism enables each spatiotemporal point in the sequence to attend to all others, allowing the network to learn the bidirectional coupling intrinsic to diffusion processes without imposing artificial causal constraints.

Through extensive evaluations, we demonstrate that HTF delivers highly accurate forward predictions near interfacial regions and, crucially, enables robust non-destructive inverse estimation of the chip's effective thermal properties using only substrate-side transient temperature data.

Compared with classical PINNs and numerical inversion baselines, HTF achieves higher accuracy, improved stability in ill-posed regimes, and enhanced physical interpretability, offering a principled and practical framework for multilayer thermal analysis in advanced semiconductor packaging.

## 2. Methodology

### 2.1 Physical Model and Governing Equations

**Fig. 1** presents the physical configuration used in this work and its one-dimensional abstraction. The system is represented by a finite thermal layer attached to a semi-infinite substrate, forming a single material interface where the thermal properties change abruptly. This reduced layer and substrate model retains the essential interfacial structure that governs transient heat conduction and serves as the basis for the mathematical formulation. The governing equations for the layer and substrate model with a single interface are given by:

$$\frac{\partial u(x,t)}{\partial t} = \alpha_S \frac{\partial^2 u(x,t)}{\partial x^2}, \quad (0 \leq x \leq L) \tag{1}$$

$$\frac{\partial u(x,t)}{\partial t} = \alpha_R \frac{\partial^2 u(x,t)}{\partial x^2}, \quad (x \geq L) \tag{2}$$

$$\kappa_S \frac{\partial u(L^-,t)}{\partial x} = \kappa_R \frac{\partial u(L^+,t)}{\partial x}, \tag{3}$$

$$u(L^-,t) = u(L^+,t) \tag{4}$$

These equations define a piecewise diffusion problem in which **Eqs. (1-2)** describe transient heat conduction within the layer and substrate, respectively, while **Eqs. (3-4)** impose perfect thermal contact through heat-flux conservation and temperature continuity at the interface. This formulation captures the essential physics of heterogeneous media, where the discontinuity in thermal diffusivity and conductivity generates sharp temperature gradients in the vicinity of the interface.

The variable $u(x,t)$ represents the temperature field, with the layer and substrate distinguished implicitly through their respective spatial domains. The coordinate $x$ measures distance normal to the interface, $t$ represents time, and L marks the location of the material interface. The symbols $\alpha_S$ and $\alpha_R$ denote the thermal diffusivities of the layer and substrate, respectively, while $\kappa_S$ and $\kappa_R$ are their corresponding thermal conductivities. The superscripts $L^-$ and $L^+$ indicate limits approaching the interface from the layer side and the substrate side, ensuring the continuity of temperature and conservation of heat flux under perfect thermal contact.

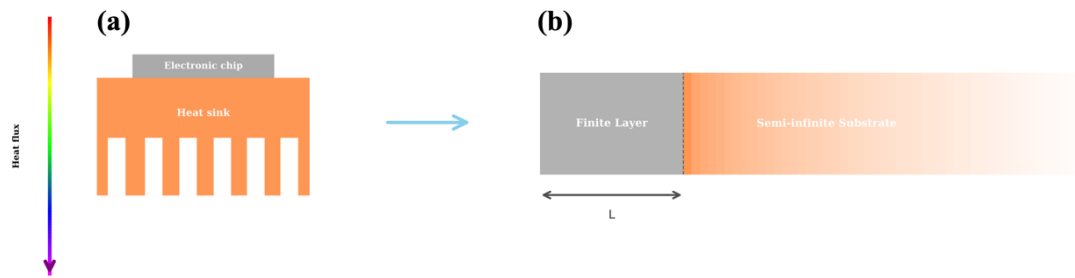

**Figure 1.** Abstraction of a practical thermal management system into a benchmark heat conduction model. (a) Heat flows from an electronic chip into a heat sink in typical electronic packaging. (b) The configuration is abstracted as a finite layer on a semi-infinite substrate, representing a one-

dimensional system with a single material interface.

To facilitate analysis and simplify the subsequent model development, the governing equations are recast in dimensionless form using the finite layer thickness and the thermal properties of the substrate as characteristic scales. We begin by normalizing the spatial coordinate and material properties through:

$$\bar{x} = \frac{x}{L}, \qquad \bar{\alpha} = \frac{\alpha_S}{\alpha_R}, \qquad \bar{K} = \frac{\alpha_S}{\alpha_R} \tag{5}$$

Substituting $\bar{x}$ into the dimensional heat equations and requiring the resulting equations to retain their canonical diffusion form yields the dimensionless time variable: $\bar{t} = \frac{\alpha_S t}{L^2}$.

With these definitions, heat conduction in the finite layer $(0 \leq \bar{x} \leq 1)$ is governed by:

$$\frac{\partial \bar{u}}{\partial \bar{t}} = \bar{\alpha} \frac{\partial^2 \bar{u}}{\partial \bar{x}^2}, \tag{6}$$

while the substrate $(\bar{x} \geq 1)$ satisfies:

$$\frac{\partial \bar{u}}{\partial \bar{t}} = \frac{\partial^2 \bar{u}}{\partial \bar{x}^2}, \tag{7}$$

Perfect thermal contact at $\bar{x} = 1$ imposes heat flux continuity:

$$\bar{K} \frac{\partial \bar{u}(1^-, \bar{t})}{\partial \bar{x}} = \frac{\partial \bar{u}(1^+, \bar{t})}{\partial \bar{x}}, \tag{8}$$

together with temperature continuity:

$$\bar{u}(1^-, \bar{t}) = \bar{u}(1^+, \bar{t}) \tag{9}$$

Here $\bar{u}$ denotes the dimensionless temperature field, and the superscripts $1^-$ and $1^+$ indicate limits approaching the interface from each side.

**2.2 Spatiotemporal Sampling Strategy**

Accurately resolving heat conduction across heterogeneous media requires capturing sharp spatiotemporal variations induced by interfacial thermal mismatches. In particular, temperature gradients and heat flux discontinuities near the interface dominate the early-time transient behavior and are extremely localized. Standard grid-based numerical methods (e.g., FDM/FEM) employ uniform discretization [7-8], while operator-learning approaches [36-37] and PINN-based adaptive samplers [38-40] typically emphasize global residual distribution. However, none of these strategies explicitly prioritize the interface-driven correlations where the physics is most expressive.

To address this, we use a spatiotemporal neighborhood extension strategy that first selects a set of anchor points across the domain and then expands them into localized spatiotemporal neighborhoods for training, thereby enhancing resolution in critical regions while preserving global thermal consistency. As illustrated in **Fig. 2(a)**, anchor points are sampled at representative spatiotemporal locations. In **Fig. 2(b)**, each anchor point is expanded into a rectangular neighborhood of size $K \times T$ in space and time, producing dense local sampling sequences. These sequences are designed to capture both spatial and temporal dependencies, with particular emphasis on regions near interfaces.

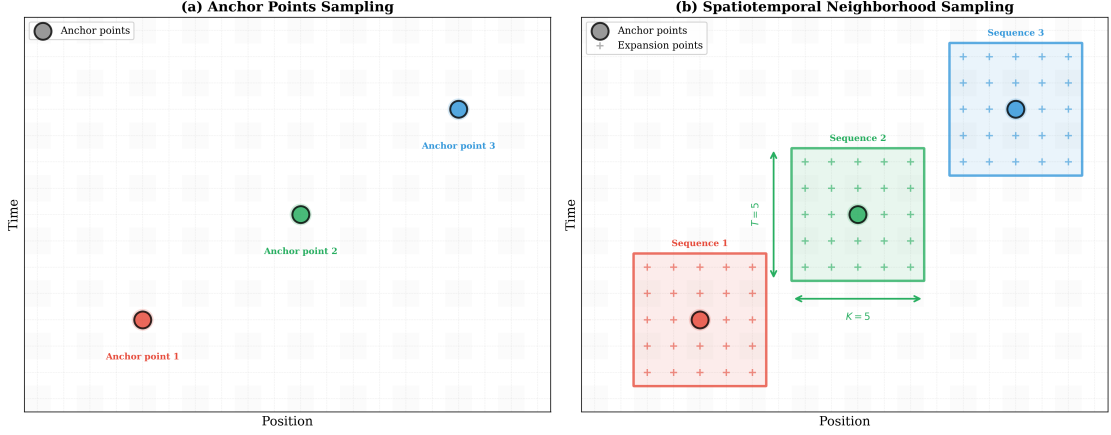

**Figure 2.** Spatiotemporal sampling strategy. (a) Anchor points sampling: sparse anchor points are selected in the spatiotemporal domain. (b) Spatiotemporal neighborhood sampling: for each anchor point, a local spatiotemporal neighborhood is sampled, defined by spatial range $K$ and temporal range $T$, to capture local temperature variations.

Specifically, we distribute a sparse set of anchor positions $\{x_1, x_2, \ldots, x_S\}$ across the spatial domain. Around each anchor point $x_i$, instead of using a single temporal sequence, we define a local spatiotemporal sequence:

$$\text{sequence}_i = \{(x, t) \mid x \in \mathcal{X}(x_i), t \in \mathcal{T}(t_i)\} \quad (10)$$

where $\mathcal{X}(x_i)$ is a small neighborhood of $x_i$, such as $\{x_i - \Delta x, x_i, x_i + \Delta x\}$ and $\mathcal{T}(t_i)$ is a small neighborhood of $t_i$, such as $\{t_0, t_0 + \Delta t, \ldots, t_0 + (T-1)\Delta t\}$. Each sequence therefore forms a mini grid of size

($K \times T$), where $K$ is the number of spatial samples and $T$ is the number of temporal steps. This sequence-level sampling provides the model with local spatial context reflecting heat diffusion around the anchor point, enabling the model to capture both spatial gradients and temporal dynamics simultaneously. The set of all such sequences across anchors forms the training dataset:

$$\mathcal{D} = \{\text{sequence}_1, \text{sequence}_2, \dots, \text{sequence}_s\} \qquad (11)$$

which is later encoded and fed into the HTF network. This design explicitly captures the fundamental nature of thermal propagation as a diffusion process evolving across space and time, while maintaining computational tractability and modeling flexibility.

## 2.3 Derivation of LaplaceAct

To enhance the physical interpretability of our model, we seek to design an activation function that aligns more closely with the analytical structure of PDE solutions, particularly those encountered in thermal conduction problems. Standard activation functions, such as ReLU [41], Tanh [42], and GELU [43-44], lack a direct correspondence with the underlying solution structures of PDEs. Recent efforts have explored physics-aware activation functions, including sinusoidal activations [45] for capturing periodic behaviors and complementary error functions [46] that appear naturally in diffusion processes. However, these approaches either focus on specific solution characteristics (periodicity or spatial profiles) or fail to offer a unified representation of the combined exponential decay and oscillatory patterns characteristic of thermal diffusion [47-48]. In contrast, the proposed LaplaceAct activation is derived directly from the inverse Laplace transform, which is widely used for solving heat conduction equations, and provides unified mathematical framework that inherently captures both transient thermal decay and spatial oscillations in heat conduction solutions.

The inverse Laplace transform of a function $F(s)$ is given by the Bromwich integral:

$$f(t) = \frac{1}{2\pi i}\int_{\gamma-i\infty}^{\gamma+i\infty} F(s)e^{st} ds, \text{where } s = -\lambda + i\omega, \lambda > 0 \qquad (12)$$

In our approach, we choose the path $s = -\lambda + i\omega$ for the Laplace inverse transform, where $\lambda > 0$ is a positive real part and $\omega$ represents the frequency. The main reason for this choice is to guarantee the presence of dissipative decay in the solution, which is a universal feature of all stable physical systems, including thermal diffusion, viscous flows, electromagnetic attenuation, and quantum decoherence [49-51]. This path choice ensures thermal dissipation and prevents unbounded growth, yielding stable and physically consistent solutions. Following this path selection, substituting the decomposition of $s$ into the integral and separating the real and imaginary parts yields:

$$f(t) = \int_{-\infty}^{+\infty} A(\omega)e^{-\lambda t}\cos(\omega t) + B(\omega)e^{-\lambda t}\sin(\omega t)d\omega \tag{13}$$

where $A(\omega)$ and $B(\omega)$ are determined by the real and imaginary parts of $F(s)$. To make this representation amenable to neural network implementation, we approximate the integral using a Riemann sum:

$$f(t) \approx \sum_{n=1}^{N} A_n e^{-\lambda t}\cos(\omega_n t) + B_n e^{-\lambda t}\sin(\omega_n t) \tag{14}$$

which motivates the construction of a corresponding activation function that reflects this structure. We then define the LaplaceAct as follows:

$$\text{LaplaceAct}(z) = \omega_1 e^{-\lambda t}\cos(\omega t) + \omega_2 e^{-\lambda t}\sin(\omega t) \tag{15}$$

where $\omega_1$, $\omega_2$, $\lambda$, $\omega$ are trainable parameters, and $t = |z|$. This activation function allows the neural network to learn combinations of decaying oscillatory modes, enabling it to approximate a wide class of physically relevant functions, particularly those arising in heat transport problems.

Then, consider a spatiotemporal coordinate $(x, t)$ mapped through a learnable embedding layer to an *n*-dimensional latent vector:

$$\mathcal{Z} = \emptyset(x, t) \in \mathbb{R}^n \tag{16}$$

Each component $z_i$ of $\mathcal{Z}$ is then passed through a shared Laplace-inspired activation function:

$$\hbar_i = \text{LaplaceAct}(z_i) = \omega_1 e^{-\lambda|z_i|}\cos(\omega|z_i|) + \omega_2 e^{-\lambda|z_i|}\sin(\omega|z_i|) \tag{17}$$

The resulting vector $\mathcal{H} = [\hbar_1, \hbar_2, \ldots, \hbar_n]$ is subsequently linearly combined via an output layer:

$$\hat{f}(x,t) = \sum_{i=1}^{n} \alpha_i \hbar_i = \sum_{i=1}^{n} \alpha_i \left[ \omega_1 e^{-\lambda|z_i|} \cos(\omega|z_i|) + \omega_2 e^{-\lambda|z_i|} \sin(\omega|z_i|) \right] \quad (18)$$

The expression is structurally equivalent to a Riemann sum approximation of the inverse transform. The LaplaceAct function thus embeds inductive bias directly derived from the structure of inverse Laplace solutions into the neural architecture, enhancing its capacity to capture both transient and steady-state features in PDE-governed systems. In the following sections, we demonstrate that networks equipped with LaplaceAct not only retain the universal approximation property but also offer improved accuracy in modeling physically consistent thermal fields.

### 2.4 HTF Model Architecture

As shown in **Fig. 3**, the HTF model leverages the LaplaceAct activation function introduced in **Section 2.2** to learn the spatiotemporal characteristics of heat transfer problems governed by PDEs. Given an input sequence of $(x, t)$ pairs, HTF first performs a linear embedding that maps the two-dimensional coordinates into a higher-dimensional latent space of size $d_{\text{model}}$. This embedded sequence is then processed through a stack of $N$ Transformer decoder blocks, each composed of a multi-head self-attention layer followed by a feed-forward subnetwork, within which the proposed LaplaceAct activation function is employed.

In each decoder block, LaplaceAct is integrated throughout both the attention and feed-forward components, with residual connections

facilitating stable training. Formally, each decoder layer updates the input hidden state via:

$$\acute{z} = z + \text{MHA}(\text{LaplaceAct}(z), \text{LaplaceAct}(z), \text{LaplaceAct}(z)) \quad (19)$$

$$z_{\text{out}} = \acute{z} + \text{FFN}(\text{LaplaceAct}(\acute{z})) \quad (20)$$

The feed-forward network (FFN) employs a three-layer architecture with LaplaceAct activations after the first two linear transformations:

$$\text{FFN}(x) = \text{Linear}_3\left(\text{LaplaceAct}\left(\text{Linear}_2\left(\text{LaplaceAct}(\text{Linear}_1(x))\right)\right)\right) \quad (21)$$

After passing through the decoder stack, the final hidden representation is processed by an output regression module consisting of two LaplaceAct-activated hidden layers and a final linear projection. This module maps the latent vector back to a single scalar field value $u(x,t)$, enabling the network to make pointwise predictions across the domain. The entire network thus implements a mapping:

$$(x, t) \mapsto u(x, t) = \text{HeatTransFormer}(x, t; \theta) \quad (22)$$

where $\theta$ includes all trainable parameters from the embedding, attention, activation, and output layers.

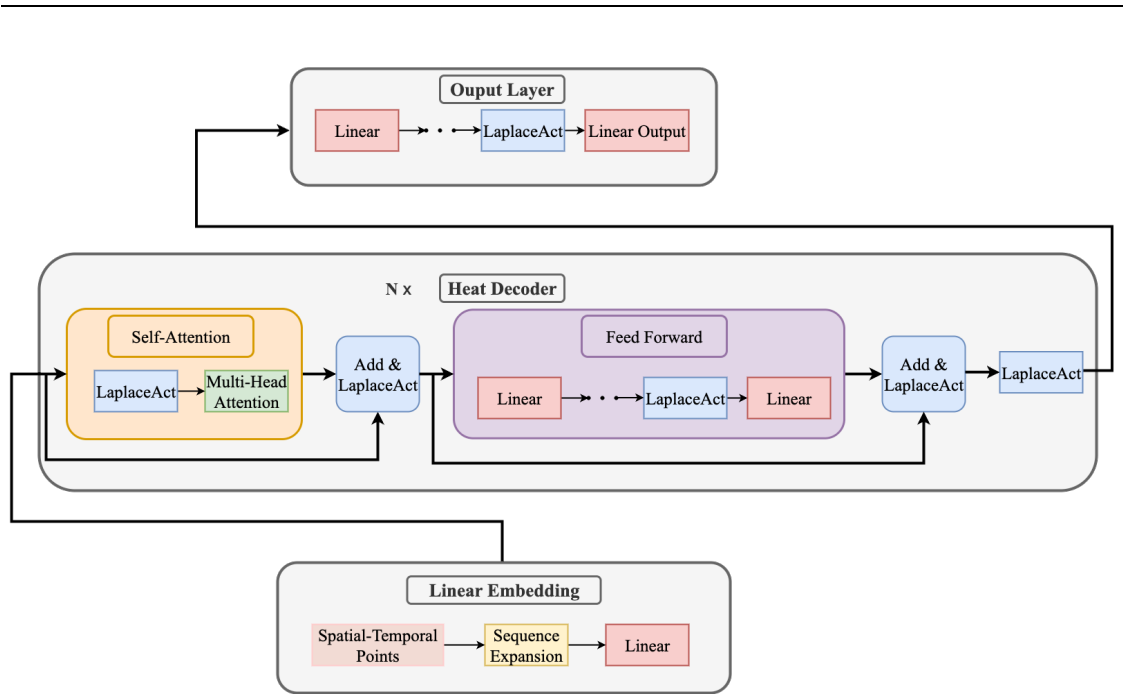

**Figure 3.** The architecture of HeatTransFormer. The model comprises: (1) Linear Embedding that transforms spatiotemporal coordinates via sequence expansion; (2) Heat Decoder with N stacked layers, each containing self-attention (orange) and feed-forward (purple) blocks with pre-activation LaplaceAct and residual connections; (3) Output Layer for temperature field regression.

## 3. Forward Problem: Benchmark Validation Against PINNs

### 3.1 Problem Formulation and Benchmark Design

To enable a fair and rigorous comparison with traditional PINNs, we design a benchmark that concentrates the types of difficulties PINNs typically encounter in multi-region transient diffusion problems [52]. Discontinuous material properties complicate the enforcement of interface conditions [53], complex boundary constraints often lead to unstable optimization [54], and sharp initial gradients introduce spectral bias that hampers the recovery of high-frequency thermal features [55-56]. By combining these effects into a single configuration, the benchmark serves as a demanding and representative test case for evaluating HTF.

Based on these considerations, we adopt the dimensionless formulation introduced in **Section 2.1**, where the substrate serves as the reference material. This benchmark simulates the contact cooling between a finite heated layer and a semi-infinite cooling substrate, a configuration commonly encountered in electronic thermal management, such as heat dissipation from semiconductor chips through thermal interface materials into heat sinks. Under this scaling, only the relative material contrasts need to be specified. The finite thermal layer is assigned a dimensionless thermal diffusivity $\bar{\alpha} = 0.3$ and a dimensionless thermal conductivity $\bar{\kappa} = 0.5$. These ratios capture the weaker heat-spreading capability of semiconductor materials relative to metallic substrates and create the diffusivity and conductivity discontinuities necessary to test the model's ability to resolve interface-driven thermal behavior. The initial condition imposes a unit temperature in the layer and zero temperature in the substrate:

$$\bar{u}(\bar{x}, 0) = \begin{cases} 1, & 0 \leq \bar{x} \leq 1, \\ 0, & \bar{x} \geq 1, \end{cases}$$

where the left boundary of the layer is adiabatic: $\frac{\partial \bar{u}(\bar{x},\bar{t})}{\partial \bar{x}}|_{\bar{x}=0} = 0$. The substrate extends semi-infinitely to the right and is truncated at $\bar{x} = 10$

with a fixed far-field temperature $\bar{u}(10, \bar{t}) = 0$.

## 3.2 Model Configuration and Training Setup

With the benchmark fully specified, the next step is to define the model architectures used for HTF and PINNs. Because the governing equations differ across the two regions, both methods adopt a two-instance setup in which each domain is handled by a separate model, and the interface conditions provide the physical coupling required to maintain temperature continuity and heat-flux balance.

For HTF, two independent Transformer instances are employed. Each instance contains one decoder layer with 32-dimensional embeddings, a hidden dimension of 512, and four attention heads. Instance 1 is responsible for predicting the temperature evolution in the finite heated layer within the range $0 \leq \bar{x} \leq 1$ and uses the layer diffusivity $\bar{\alpha} = 0.3$. Instance 2 models the semi-infinite substrate for $\bar{x} \geq 1$ and uses the reference diffusivity $\bar{\alpha} = 0.1$. During training, the two Transformer instances operate on localized spatiotemporal sequences constructed using the sampling strategy described in **Section 2.2**, and the interface conditions are imposed directly within the loss function to ensure consistency between the two domains.

The PINNs baseline follows the same two-instance architecture for a fair comparison. Two independent fully connected networks are constructed, one for each domain, and each network contains four hidden layers with 512 neurons per layer. Unlike HTF, which processes structured spatiotemporal sequences, the PINNs approach operates on individual coordinate points and relies entirely on physics-informed losses to encode the governing equations, boundary constraints, and interface conditions. All hyperparameters used for HTF and PINNs in the comparison study are summarized in **Table 1**.

| Model | Hyperparameter | Value | Trainable weight |
|---|---|---|---|
| PINNs | Hidden layer | 4 | 1054722 |
|  | Hidden neuron | 512 |  |
| HTF | Decoder | 1 | 733726 |
|  | Embedding size | 32 |  |
|  | Head | 4 |  |
|  | Hidden neuron | 512 |  |

**Table 1** Hyperparameter configurations and total number of trainable weights for PINNs and HTF models used in the benchmark comparison.

Accurate resolution in the vicinity of the material interface is essential for representing the steep thermal gradients that arise during early-time diffusion. For both approaches, the temporal domain $0 \leq \bar{t} \leq 1$ is discretized into 20 uniformly time points. For the HTF approach, to resolve the sharp gradients near the interface, we introduce 20 anchor points in the narrow region $0 \leq \bar{x} \leq 1$ adjacent to the material boundary, and 30 anchor points are distributed in the far-field region $1 \leq \bar{x} \leq 10$. After temporal and spatial expansion and following the discretization step sizes of $\Delta \bar{x} = 1 \times 10^{-3}$ and $\Delta \bar{t} = 1 \times 10^{-3}$ as specified in **Eq. (10)**, this scheme results in a total of 5000 collocation points.

For the PINNs baseline, a denser pointwise sampling scheme is adopted because the model does not benefit from the sequence-level neighborhood enrichment used by HTF. To ensure comparable resolution near the interface, 100 collocation points are assigned to $0 \leq \bar{x} \leq 1$ and 150 points to $1 \leq \bar{x} \leq 10$, for a total of 5000 points. This balanced allocation provides sufficient spatial resolution to capture the sharp gradients at the interface while maintaining consistent computational cost between the two approaches.

The training objective incorporates several physics-based constraints through a carefully weighted composite loss that balances the enforcement of the governing PDEs, the satisfaction of boundary conditions, and the continuity requirements at the material interface. The total loss function is expressed as a weighted sum of the corresponding physics-informed terms:

$$\mathcal{L}_{\text{Total}} = \lambda_{\text{PDE}_1} \mathcal{L}_{\text{PDE}_1} + \lambda_{\text{PDE}_2} \mathcal{L}_{\text{PDE}_2} + \lambda_{\text{interface}_T} \mathcal{L}_{\text{interface}_T} +$$

$$\lambda_{\text{interface}_{\text{flux}}} \mathcal{L}_{\text{interface}_{\text{flux}}} + \lambda_{\text{BC}_{\text{left}}} \mathcal{L}_{\text{BC}_{\text{left}}} + \lambda_{\text{BC}_{\text{right}}} \mathcal{L}_{\text{BC}_{\text{right}}} + \lambda_{\text{IC}_1} \mathcal{L}_{\text{IC}_1} +$$
$$\lambda_{\text{IC}_2} \mathcal{L}_{\text{IC}_2},$$

where the loss comprises: PDE residuals $\lambda_{\text{PDE}_1} \mathcal{L}_{\text{PDE}_1} + \lambda_{\text{PDE}_2} \mathcal{L}_{\text{PDE}_2}$ ensuring heat equation satisfaction in both domains, interface constraints $\lambda_{\text{interface}_T} \mathcal{L}_{\text{interface}_T} + \lambda_{\text{interface}_{\text{flux}}} \mathcal{L}_{\text{interface}_{\text{flux}}}$ enforcing temperature and heat flux continuity at the interface, boundary conditions $\lambda_{\text{BC}_{\text{left}}} \mathcal{L}_{\text{BC}_{\text{left}}} + \lambda_{\text{BC}_{\text{right}}} \mathcal{L}_{\text{BC}_{\text{right}}}$ at the left and right boundaries, and initial conditions $\lambda_{\text{IC}_1} \mathcal{L}_{\text{IC}_1} + \lambda_{\text{IC}_2} \mathcal{L}_{\text{IC}_2}$ for both regions. We assign $\lambda_{\text{interface}_T} = \lambda_{\text{interface}_{\text{flux}}} = 5$ to emphasize the critical role of interface constraints in this problem, where accurate interface coupling is essential for modeling materials with different thermal properties. All other weighting coefficients are set to unity.

Both approaches utilize the Adam optimizer with a learning rate of $1 \times 10^{-3}$ over 500 training epochs, with all experiments conducted on a single NVIDIA RTX 4090 GPU (24GB) for computational efficiency. To ensure reproducible results, all random number generators are initialized with a fixed seed value of 0. For reproducibility and to facilitate further research, all code implementations and demonstration examples are made publicly and freely available at: https://github.com/manicsetsuna/heattransformer.

### 3.3 Results and Analysis

**Fig. 4** compares the performance of HTF and PINNs on the benchmark problem defined in **Section 3.1**. HTF accurately reconstructs the interface-driven thermal dynamics and closely matches the analytical solution [57] across the entire spatiotemporal domain. The predicted temperature fields preserve the sharp transitions associated with discontinuous material properties and correctly capture the early-time boundary layer as well as the long-time diffusive behavior.

The PINNs solution, in contrast, exhibits substantial deviations from the analytical benchmark. It produces unphysical temperature undershoots and irregular spatial oscillations, indicating that the pointwise formulation fails to consistently enforce the governing equations together with the boundary and interface conditions. These artifacts highlight a fundamental limitation of PINNs when dealing with multi-region diffusion.

The quantitative error analysis further reinforces this conclusion. HTF maintains low and spatially coherent error distributions across all metrics. The pointwise $L_1$ and $L_2$ errors remain below 0.07 and 0.005, respectively, while the RMSE profile demonstrates uniform accuracy throughout the domain. PINNs, on the other hand, display significantly higher and irregular errors, especially near the interface where the diffusivity and conductivity change abruptly.

Overall, HTF reduces the error by approximately $70\% - 80\%$ compared with PINNs and delivers consistent accuracy in regions dominated by steep thermal gradients. These improvements confirm HTF's strong capability in modeling interfacial heat transfer, and additional comparisons under different boundary conditions are provided in **Fig. S1** of the Supplementary Materials.

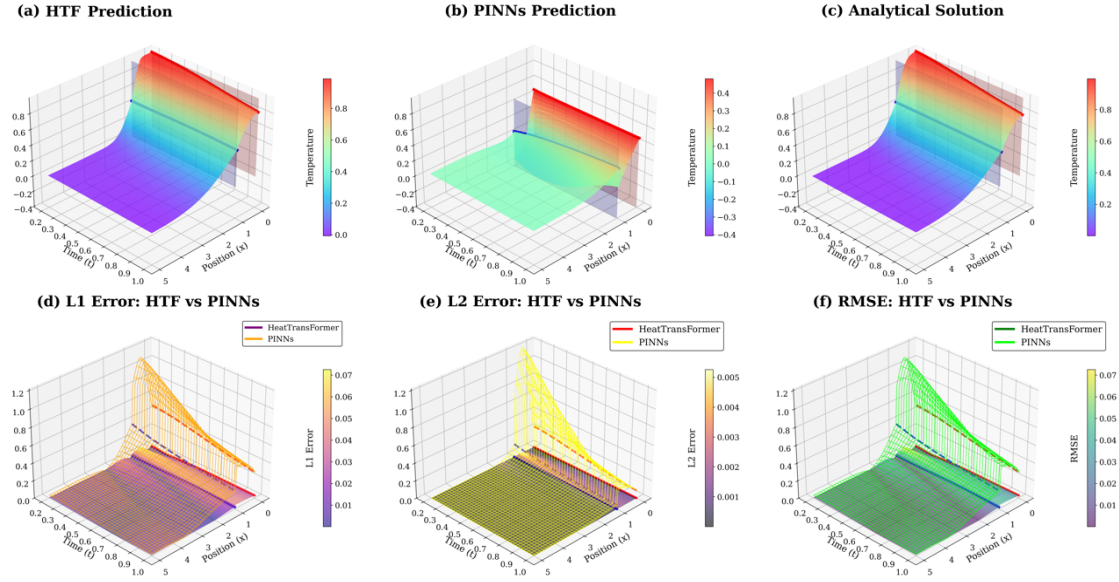

**Figure 4.** Comparison of HTF and PINNs on the benchmark problem. (a-c) Predicted temperature fields from HTF, PINNs, and the analytical solution, with the domain truncated at $\bar{x} = 5$. (d-f) Pointwise $L_1$, $L_2$, and RMSE error distributions comparing HTF (colored surfaces) and PINNs (wireframes). The red curves represent the temperature or error values extracted along the left boundary at $\bar{x} = 0$, while the blue curves represent those extracted at the material interface at $\bar{x} = 1$. These traces are included to highlight model performance in boundary- and interface-dominated regions, where accurate enforcement of physical constraints is most critical. Specifically, Pointwise $L_1$ error is computed as: $L_1(\bar{x}, \bar{t}) =$

$\left|u^{\text{pred}}(\bar{x}, \bar{t}) - u^{\text{true}}(\bar{x}, \bar{t})\right|$, and Pointwise $L_2$ error is computed as: $L_2(\bar{x}, \bar{t}) = \left(u^{\text{pred}}(\bar{x}, \bar{t}) - u^{\text{true}}(\bar{x}, \bar{t})\right)^2$.

## 4. Physical Insights into Model Performance

### 4.1 Spatiotemporal Sampling Strategy Analysis

Building on the observations in **Section 3**, we investigate why HTF achieves substantially higher accuracy than PINNs. A central hypothesis is that HTF benefits from its ability to incorporate local spatiotemporal structure through sequence extension, whereas pointwise sampling lacks such contextual information. To isolate the impact of this mechanism, we compare two HTF variants: one with sequence extension and one without it, both trained on the same layer and substrate benchmark. The geometry, governing equations, and material parameters are kept identical, while boundary and initial conditions are modified to create a new transient cooling scenario. Apart from the sampling strategy, all architectural and training components remain unchanged, ensuring a controlled comparison.

Both variants use one decoder layer, 32-dimensional embeddings, 512-dimensional hidden states, and four attention heads. Boundary conditions impose $\bar{u}(0, \bar{t}) = 1.0$ for heating and $\bar{u}(+\infty, \bar{t}) = 0$ as the ambient reference, while interface temperature and heat flux continuity are enforced at $\bar{x} = 1$. The entire domain is initialized at ambient temperature.

Training is performed using the Adam optimizer with a learning rate of $1 \times 10^{-3}$ for 500 epochs. To ensure a fair comparison between the two sampling strategies, we allocate 5000 spatiotemporal points to the sequence-extension approach and an equal number of independent points to the pointwise sampling approach, thereby maintaining comparable computational complexity across both experiments.

The results in **Fig. 5** demonstrate the essential role of spatiotemporal context in HTF. The configuration with sequence extension consistently reconstructs smooth, physically realistic temperature fields that closely follow the analytical solution. In contrast, the pointwise variant exhibits oscillatory and unphysical behavior, especially near the interface, where diffusive smoothness is most critical.

Error analyses further reinforce this trend. The global $L_1$ error of the sequence-based model remains below $1.2 \times 10^{-1}$, representing a two- to three-fold improvement over the pointwise variant. The $L_2$ errors show even larger gains, with consistent reductions by factors of three to five. Insets near the interface ($\bar{x} \in [0.8, 1.2]$) reveal that sequence extension preserves both smooth temperature transitions and physically consistent gradients, while the independent sampling approach fails to maintain these properties.

These improvements originate from the physical structure of heat diffusion. Heat conduction is inherently local, governed by the diffusion operator $\nabla^2 \bar{u}$, which requires information from neighboring spatial points, together with temporal evolution $\frac{\partial \bar{u}}{\partial \bar{t}}$, which couples the current temperature to its temporal neighborhood. As a result, accurate modeling requires access to spatiotemporal correlations rather than isolated points.

Sequence extension aligns naturally with this physical locality principle. By expanding each anchor into a neighborhood, HTF provides the attention mechanism with the necessary local context to construct meaningful feature representations. Furthermore, the self-attention mechanism dynamically allocates weights to spatial and temporal neighbors, effectively forming adaptive, data-driven stencils. Unlike fixed-coefficient schemes such as $(1, -2, 1)$ used in finite difference methods, the adaptive attention weights adjust automatically to local conditions, emphasizing relevant neighbors in regions with sharp gradients and maintaining stability in smoother regions. This adaptive also weighting explains HTF's ability to resolve steep interfacial gradients and early-time transients with high fidelity, while pointwise sampling fails to capture these physically essential couplings.

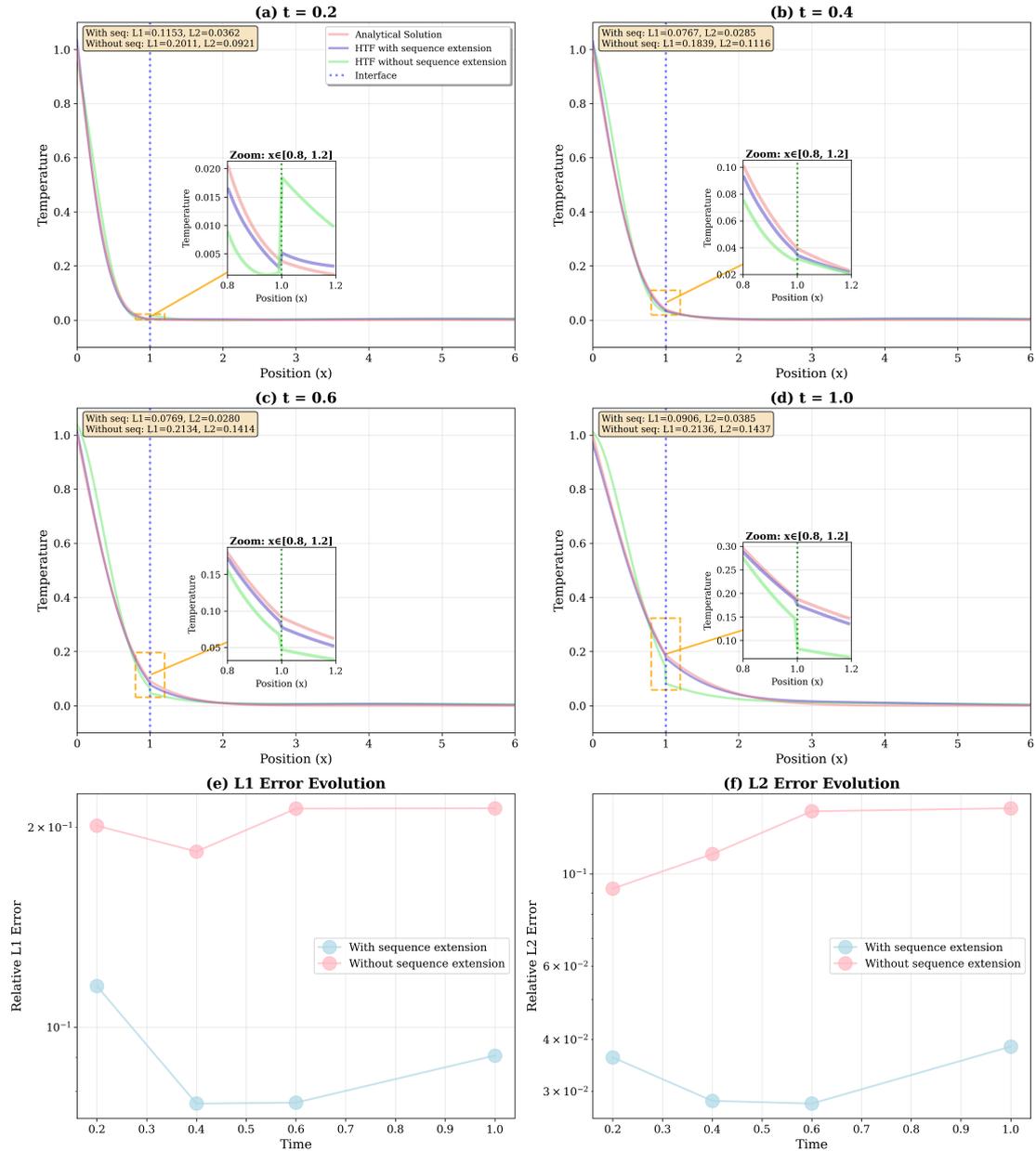

**Figure 5.** Comparison of sequence-extension strategies on the finite and semi-infinite heat conduction problem. (a-d) Predicted temperature profiles at $\bar{t} = 0.2, 0.4, 0.6,$ and $1.0$, compared with analytical solutions. Insets show zoomed views near the interface ($\bar{x} \in [0.8, 1.2]$) to highlight prediction differences. (e-f) Global $L_1$ and Global $L_2$ error evolution over time, demonstrating the superior accuracy of the sequence extension method. Log-scale plots emphasize error reduction. Specifically, Global

$L_1$ error is computed as: $L_1(\bar{x}, \bar{t}) = \frac{\sum_i |\bar{u}_i^{pred}(\bar{x},\bar{t}) - \bar{u}_i^{true}(\bar{x},\bar{t})|}{\sum_i |\bar{u}_i^{true}(\bar{x},\bar{t})|}$, and Global L2 error is computed as: $L2(\bar{x}, \bar{t}) = \frac{(\sum_i (\bar{u}_i^{pred}(\bar{x},\bar{t}) - \bar{u}_i^{true}(\bar{x},\bar{t}))^2)^{1/2}}{(\sum_i (\bar{u}_i^{true}(\bar{x},\bar{t}))^2)^{1/2}}$.

## 4.2 LaplaceAct Activation Performance Analysis

**Section 2.3** introduced LaplaceAct as a physics-informed activation function derived from the analytical structure of inverse Laplace transforms. This construction naturally encodes the exponential relaxation and damped oscillatory signatures that define transient heat conduction. To examine whether this theoretical advantage translates into practical performance gains, we conduct a controlled comparison against several commonly used activation functions, using the same benchmark as in **Section 4.1** and keeping all physical settings, while keeping all physical settings, model architectures, and training parameters identical.

In this experiment, we compare five activation functions: LaplaceAct, ReLUAct [41], TanhAct [42], ErfcAct [46], and SinAct [58], using identical architectures and training configurations. For completeness, the explicit mathematical forms of all activation functions are provided in the Supplementary Materials. While ErfcAct inherits part of its structure from classical diffusion solutions, the remaining three represent standard nonlinearities widely used in scientific machine learning. All experiments adopt the same anchor-based spatiotemporal sampling strategy detailed in **Section 4.1** and are trained for 500 epochs using Adam with a learning rate of $10^{-3}$, ensuring that any performance differences arise solely from the activation function.

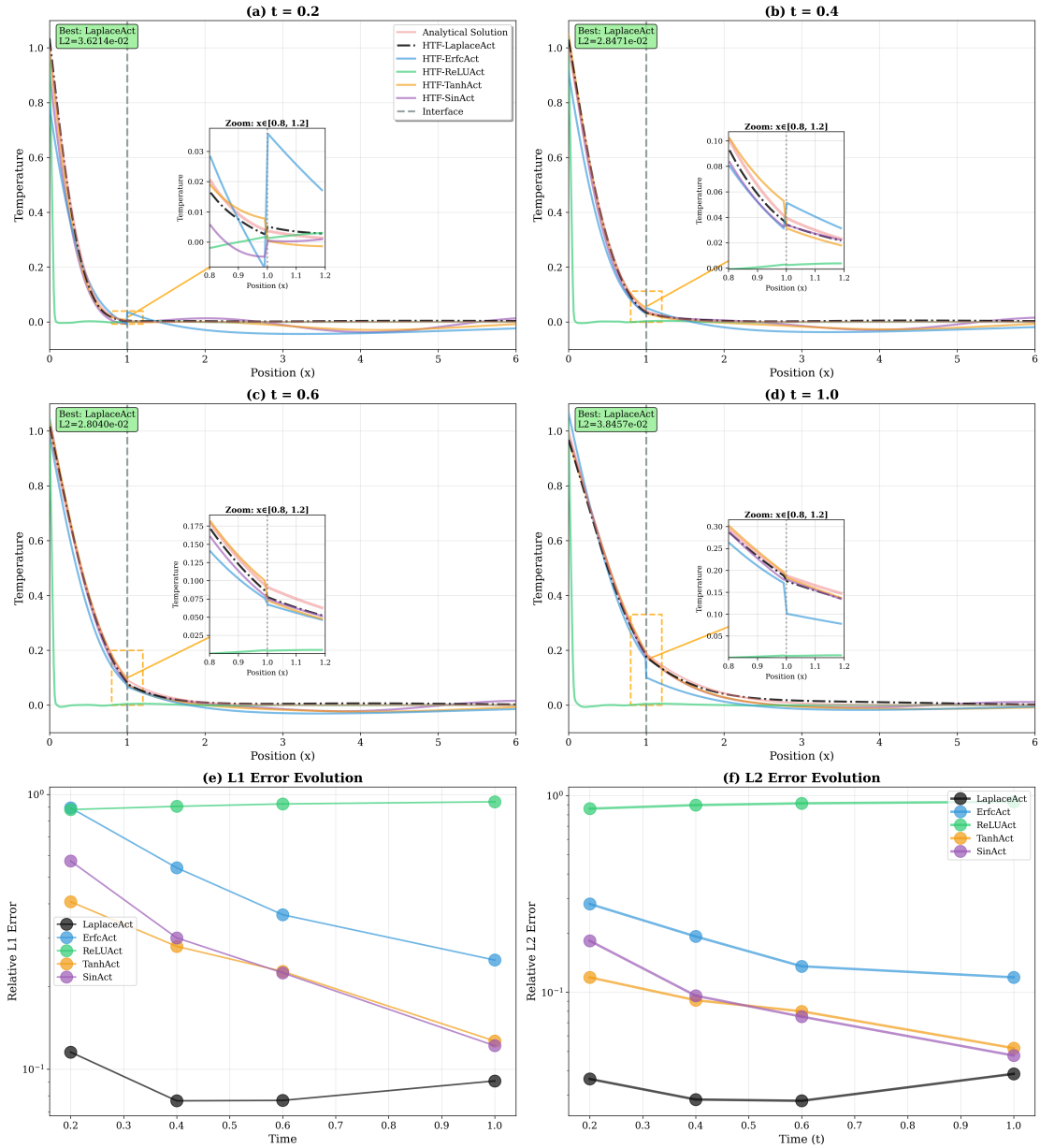

**Figure 6.** Comparison of HTF with different activation functions on the finite and semi-infinite heat conduction problem. (a-d) Predicted temperature profiles compared with analytical solutions. Insets show detailed views near the interface ($x \in [0.8, 1.2]$). (e-f) Global $L_1$ and $L_2$ error evolution over time, confirming the robustness and accuracy of

LaplaceAct. Note: The error calculation method in **Figure 6** is consistent with that in **Figure 5**.

The predicted temperature fields exhibit markedly different behaviors, as shown in **Fig. 6(a-d)**. LaplaceAct closely tracks the analytical solution across all temporal regimes, accurately capturing both the steep early-time gradients and the smooth late-time decay. In contrast, conventional activations tend to oversmooth the interface or generate spurious oscillations, which are particularly evident in the rapid early transient period. The zoomed views around the interface further illustrate that LaplaceAct preserves smooth and physically consistent temperature gradients without exhibiting numerical artifacts. These qualitative observations are fully consistent with the quantitative error analysis, as shown in **Fig. 6(e-f)**. LaplaceAct maintains uniformly low relative $L_1$ and $L_2$ errors, typically an order of magnitude smaller than those obtained with conventional activations. Importantly, the improvement persists across both transient and quasi-steady regimes, indicating that the benefits of LaplaceAct are not tied to a specific time window but reflect a consistent enhancement across the entire diffusion process.

These observations naturally raise the question of why LaplaceAct consistently outperforms conventional nonlinearities. The origin of this performance gap lies in the inductive bias encoded by LaplaceAct. Heat conduction solutions exhibit exponential decay, interface-induced curvature, and time-evolving spatial scales, structures that LaplaceAct approximates natively due to its derivation from inverse Laplace transforms. Traditional activations lack these physics-aligned features: ReLUAct cannot express smooth curvature, TanhAct saturates prematurely, SinAct imposes artificial periodicity, and ErfcAct captures only part of the necessary solution structure. In contrast, LaplaceAct aligns naturally with the mathematical form of the diffusion solution, enabling superior accuracy, stability, and physical consistency.

### 4.3 Physical Interpretability Analysis

The analyses in **Sections 4.1** and **4.2** motivate a closer examination of HTF's interpretability. HTF provides multi-level physical interpretability spanning from network architecture to mathematical representation, creating a direct correspondence between computational processes and physical phenomena.

At the architecture level, the network workflow is designed to correspond to the physical mechanisms of heat conduction. The embedding layer performs thermal fingerprint encoding, transforming raw spatiotemporal coordinates into high-dimensional representations that capture thermal properties, boundary conditions, and historical evolution. Each component then undergoes LaplaceAct activation, which decomposes the encoded information into fundamental thermal modes. Residual connections propagate these modes across layers, ensuring stable and physically consistent evolution of the learned representation.

At the mathematical structure level, the network output aligns directly with analytical solution forms, expressing the temperature field as physically interpretable combinations of exponential decay and oscillatory modes. Successive linear transformations and activations emulate the stepwise evolution of thermal processes, enabling mode interactions while preserving their physical character. This design ensures that the network reproduces the governing physics instead of merely fitting numerical trends.

## 5. Inverse Problem: Identifying Inaccessible Thermal Properties

### 5.1 Motivation and Problem Statement

Building on the forward modeling results in **Section 3** and the interpretability analysis in **Section 4**, we now consider a significantly more challenging task: identifying unknown material properties from temperature observations. This inverse characterization problem is central to nondestructive thermal diagnostics, where parameters such as thermal diffusivity and conductivity must be inferred without directly accessing the specimen.

To assess HTF in this setting, we formulate an inverse problem based on the dimensionless finite layer and semi-infinite substrate model introduced in **Section 2.1**. As illustrated in **Fig. 7**, sample represents a finite layer occupying $0 \leq \bar{x} \leq 1$ with unknown dimensionless thermal properties $(\bar{\alpha}, \bar{\kappa})$. Reference material represents a semi-infinite substrate occupying $\bar{x} \geq 1$, whose dimensionless material properties have been normalized to unity according to the formulation in **Section 2.1**. The material interface is located at $\bar{x} = 1$, where temperature and heat flux continuity are enforced according to **Eqs. (7-8)** under the assumption of perfect thermal contact. A prescribed temperature is imposed at the accessible boundary of the finite layer, while the far end of the substrate is maintained at ambient temperature, forming a classical heating-cooling configuration commonly used in nondestructive thermal evaluation. In practice, direct instrumentation of the internal layer is often infeasible, so temperature sensors are mounted only on the outer surface of the substrate, and the internal properties must be inferred indirectly.

In the numerical experiments, measured data are generated by solving the forward problem with a high-fidelity finite element solver using known ground-truth values of $(\bar{\alpha}, \bar{\kappa})$. The resulting temperature field on the substrate surface is sampled at multiple spatial positions $\bar{x}_i$ and time instances $\bar{t}_j$ to obtain reference data $\bar{u}_{\text{measured}}(\bar{x}_i, \bar{t}_j)$, which are then corrupted with additive noise to emulate realistic sensor uncertainty. The HTF inverse solver treats $(\bar{\alpha}, \bar{\kappa})$ as trainable parameters and estimates them by minimizing the discrepancy between the measured and predicted substrate-surface temperatures, thereby recovering the unknown material properties within the framework of the finite layer and semi-infinite substrate model.

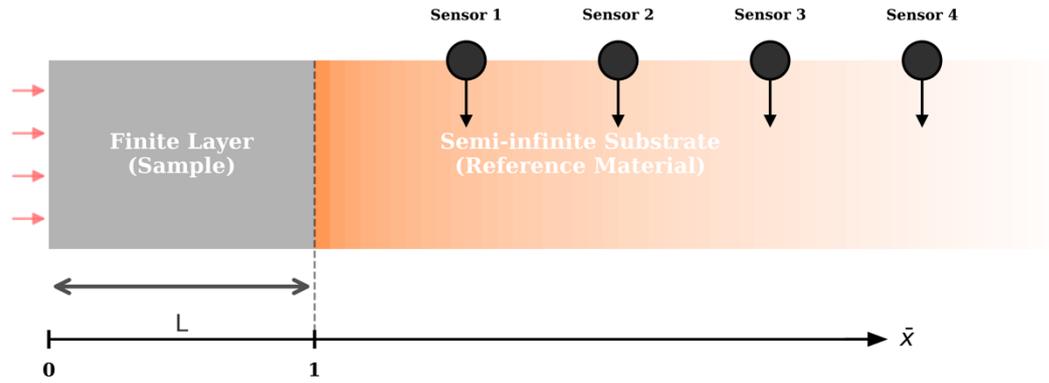

**Figure 7.** Schematic of the virtual inverse-characterization experiment. A finite layer with unknown dimensionless thermal properties $(\bar{\alpha}, \bar{\kappa})$ is placed in perfect thermal contact with a semi-infinite reference substrate whose properties are normalized to unity under the dimensionless formulation of **Section 2.1**. A constant-temperature boundary is applied at the left end of the finite layer, while the far field of the substrate is held at ambient temperature. Temperature sensors $(\bar{x} > 1)$ are mounted on the substrate surface and record time-resolved measurements $\bar{u}_{\text{measured}}(\bar{x}_i, \bar{t}_j)$.

Despite this measurement strategy, the inverse problem remains severely ill-posed. The simultaneous recovery of two thermally coupled parameters (which both influence temperature evolution) further magnifies this instability, introducing strong non-convexity and non-uniqueness into the optimization landscape. To address these challenges, we develop a physics-constrained optimization framework detailed in **Section 5.2**.

**5.2 Material Property Constraints**

Small perturbations in temperature measurements can translate into disproportionately large uncertainties in the inferred parameters. A physically meaningful solution therefore requires embedding prior knowledge about admissible material behavior directly into the parameterization of the inverse problem. In this work, this physical structure emerges naturally from the dimensionless finite layer and semi-infinite substrate formulation introduced in **Section 2.1**, which yields a compact and well-normalized representation of the thermal parameters. Crucially, the normalization step ties the inverse problem to a reference-based physical scale, compressing the admissible parameter domain and suppressing the otherwise severe drift of diffusivity and conductivity pairs into non-physical regions.

Such a reference scale is only meaningful if the underlying material properties are stable and well characterized. For this reason, we choose copper as the reference substrate, owing to its well-characterized and stable thermophysical properties widely used in thermal metrology. Once copper is adopted as the reference material, the dimensionless formulation follows naturally: $\bar{\alpha} = \frac{\alpha}{\alpha_{Cu}}$, $\bar{\kappa} = \frac{\kappa}{\kappa_{Cu}}$, $\bar{\rho}\bar{c} = \frac{\bar{\kappa}}{\bar{\alpha}}$, where copper has $\alpha_{Cu} = 1.14 \times 10^{-4}$ m²/s, $\kappa_{Cu} = 398$ W/(m·K), and $\rho_{Cu} = 3.49 \times 10^6$ J/(m³·K). This normalization not only preserves the mathematical structure of the governing equations, but also restricts the admissible parameter space to a physically meaningful region in the diffusivity and conductivity domain.

Within this domain, the dimensionless properties satisfy the thermodynamic relation $\log(\bar{\alpha}) = \log(\bar{\kappa}) - \log(\bar{\rho}\bar{c})$, or equivalently $\bar{\alpha} = \bar{\kappa}/\bar{\rho}\bar{c}$, which generates families of constant diffusivity lines. As shown in **Fig. 8**, these lines organize the Ashby map into distinct material clusters: metals and ceramics lie along high diffusivity bands, whereas polymers and natural materials populate lower diffusivity regions. These structural patterns provide a powerful physical prior for the inverse problem.

Given these material clusters, our study naturally concentrates on the subset of high conductivity and high diffusivity materials, which are most relevant to semiconductor packaging and electronic thermal management. Accordingly, we restrict the inverse search to the region these materials predominantly occupy in the dimensionless property space. In dimensionless form, this corresponds approximately to $\bar{\alpha} \in [0.1, 0.9]$, $\bar{\kappa} \in [0.01, 0.9]$ with feasible volumetric heat capacity bounded by $\bar{\rho}\bar{c} \in [0.04, 1.3]$. These constraints are enforced via hard clamping on $\bar{\alpha}$ and $\bar{\kappa}$, complemented by a soft regularization on $\bar{\rho}\bar{c}$. This strategy prohibits excursions into nonphysical regions, stabilizes the inverse optimization, and ensures that all recovered properties remain consistent with the thermal behavior of real materials.

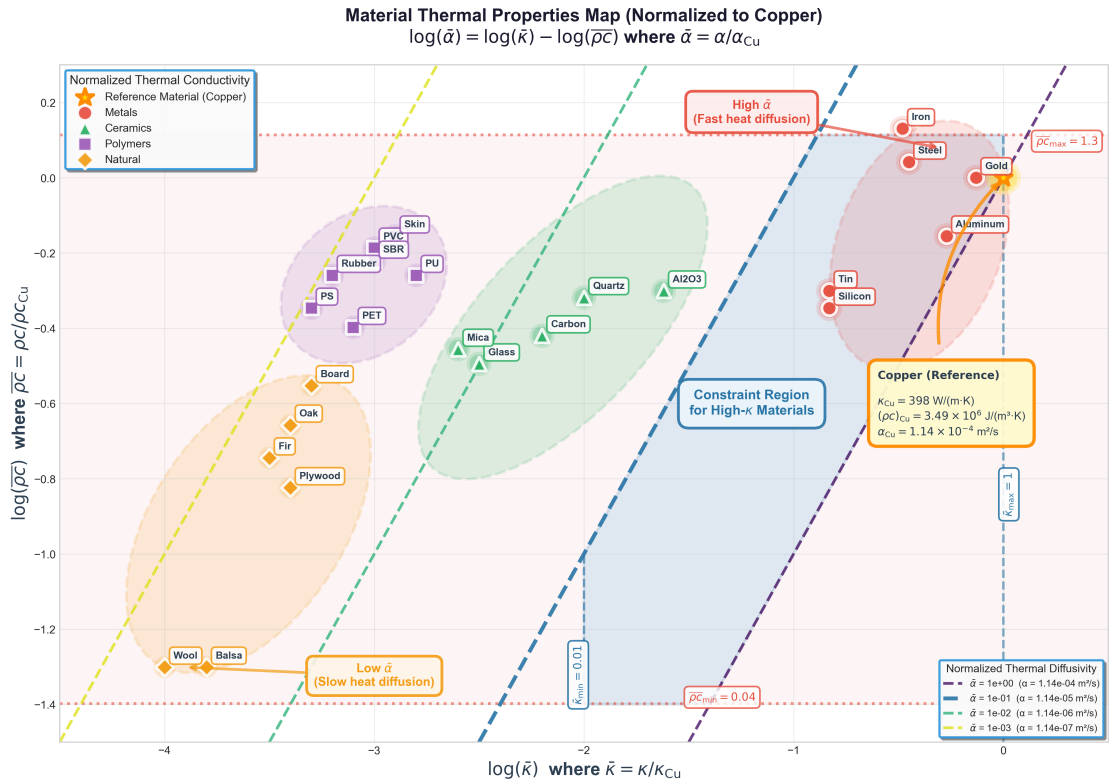

**Figure 8.** Ashby-style map of normalized thermal properties for various materials. Properties [59] are normalized to copper (reference material, marked by gold star). The parallel dashed lines represent constant thermal diffusivity values ($\bar{\alpha} = \alpha/\alpha_{Cu}$). Materials are grouped into metals (red), ceramics (green), polymers (purple), and natural materials (orange), with ellipses highlighting each category's distribution. The horizontal dotted lines indicate the typical range of normalized volumetric heat capacity ($0.04 \leq \bar{\rho}\bar{c} \leq 1.3$) for engineering materials. The blue shaded region highlights the parameter space explored in this work ($\bar{\alpha} \in [0.1, 1.0]$, $\bar{\kappa} \in [0.01, 1.0]$), targeting high-conductivity materials. Copper, serving as the reference material, is marked by a gold star at $\bar{\alpha} = \bar{\kappa} = 1.0$.

## 5.3 Synthetic Data Generation and Two-stage Optimization

To systematically evaluate HTF's inverse identification capability, we design a comprehensive numerical experiment that combines controlled synthetic data generation with a physics-informed two-stage optimization strategy. The numerical experiment begins by generating ground truth measurements through forward simulation. We solve the coupled heat conduction problem using FEA with known true parameters: $\bar{\alpha} = 0.2$ and $\bar{\kappa} = 0.24$ for the sample region. Boundary conditions are set to $\bar{u}(0, \bar{t}) = 1.0$ at the sample's left end and $\bar{u}(\infty, \bar{t}) = 0$ at the reference material's far boundary, with both regions initialized at zero temperature. Virtual temperature sensors are then placed at six strategic locations within the reference material region: $\bar{x}_{sensors} = [1.05, 1.1, 1.5, 2.0, 3.0, 4.0]$. This placement strategy concentrates sensors near the interface ($\bar{x}_{sensors} = 1.05, 1.1$) to capture the subtle heat flux signature governed by $\bar{\kappa}$, while maintaining adequate coverage in the mid-field ($\bar{x}_{sensors} = 1.5, 2.0$) to resolve the thermal diffusion dynamics controlled by $\bar{\alpha}$, and sparse sampling in the far-field ($\bar{x}_{sensors} = 3.0, 4.0$) for boundary constraint enforcement. At each sensor location, temperature is recorded at 20 uniformly spaced time steps over the interval $[0, 1.0]$, yielding a total of 120 measurement points. To simulate realistic sensor uncertainty, we corrupt these measurements with additive Gaussian noise: $\bar{u}_{\text{measured}}(\bar{x}_i, \bar{t}_j) = \bar{u}_{\text{True}}(\bar{x}_i, \bar{t}_j) + \varepsilon$, where $\varepsilon \sim N(0, \sigma^2)$, with $\sigma = 1\%$.

The inverse optimization employs a two-stage training strategy built upon the two-instance architecture introduced in **Section 3.2**. In this architecture, Instance 1 models the finite sample region, while Instance 2 models the semi-infinite reference substrate, with both instances sharing the same hyperparameter configurations as in the forward benchmark. In Stage A, we train Instance 2 (representing the reference material with known properties) using the measured sensor data with known thermal parameters to reconstruct the complete temperature field in the reference material, while keeping Instance 1 frozen. The objective loss function is:

$$\mathcal{L}_A = \lambda_{\text{PDE}_2}\mathcal{L}_{\text{PDE}_2} + \lambda_{\text{BC}_2}\mathcal{L}_{\text{BC}_2} + \lambda_{\text{IC}_2}\mathcal{L}_{\text{IC}_2} + \lambda\mathcal{L}_{\text{sensor}} + \lambda_{\text{reg}_2}\mathcal{L}_{\text{reg}_2}$$

where $\mathcal{L}_{\text{sensor}} = \|\bar{u}_{\text{HTF}} - \bar{u}_{\text{measured}}\|^2$ enforces agreement with experimental measurements. Crucially, although sensors are not placed at the interface, the PDE constraints enable Instance 2 to extrapolate the temperature field to the interface location ($\bar{x} = 1$), effectively providing $\bar{u}(1^+, \bar{t})$ as a data-driven boundary condition for Stage B. This stage essentially solves an inverse problem in the reference region: inferring the complete spatial-temporal temperature distribution from sparse sensor measurements.

With Instance 2 fixed in Stage B, the interface temperature $\bar{u}(1^+, \bar{t})$

now acts as a virtual boundary condition for the sample region. We train Instance 1 and simultaneously optimize the unknown parameters ($\bar{\alpha}, \bar{\kappa}$) by enforcing interface conditions. The interface provides two independent physical constraints: temperature continuity and heat flux conservation. The objective function becomes:

$$\mathcal{L}_B = \lambda_{\text{PDE}_1}\mathcal{L}_{\text{PDE}_1} + \lambda_{\text{BC}_1}\mathcal{L}_{\text{BC}_1} + \lambda_{\text{IC}_1}\mathcal{L}_{\text{IC}_1} + \lambda_{\text{interface}_T}\mathcal{L}_{\text{interface}_T}$$
$$+ \lambda_{\text{interface}_{\text{flux}}}\mathcal{L}_{\text{interface}_{\text{flux}}} + \lambda_{\bar{\rho}\bar{c}}\mathcal{L}_{\bar{\rho}\bar{c}} + \lambda_{\text{reg}_1}\mathcal{L}_{\text{reg}_1}$$

where $\mathcal{L}_{\text{interface}_T} = \|\bar{u}(1^-, \bar{t}) - \bar{u}(1^+, \bar{t})\|^2$ and $\mathcal{L}_{\text{interface}_{\text{flux}}} = \left\|\bar{\kappa}\frac{\partial \bar{u}(1^-, \bar{t})}{\partial \bar{x}} - \frac{\partial \bar{u}(1^+, \bar{t})}{\partial \bar{x}}\right\|^2$ enforce temperature continuity and heat flux conservation and $\mathcal{L}_{\bar{\rho}\bar{c}}$ provides a soft constraint on the volumetric heat capacity ratio to ensure thermodynamic consistency,

The heat flux continuity condition is particularly effective for identifying $\bar{\kappa}$, as it directly couples the unknown thermal conductivity with the known temperature gradient from the fixed Instance 2. Concurrently, the thermal diffusivity $\bar{\alpha}$ is constrained through the parabolic PDE in the sample, which must satisfy both the governing physics and compatibility with the interface conditions imposed by the fixed Instance 2.

This sequential training strategy effectively decouples the optimization challenge: Stage A establishes accurate interface boundary conditions using measured data, while Stage B exploits the physics-based coupling at the material interface to inversely identify the unknown sample properties without requiring direct measurements in the inaccessible region. The two-stage approach transforms the ill-posed inverse problem into a well-constrained optimization by leveraging the physical continuity laws to propagate experimental information from the accessible reference region to the unknown sample parameters.

### 5.4 Numerical Results and Performance Analysis

Our synthetic test case with $\bar{\alpha} = 0.2$ and $\bar{\kappa} = 0.24$ corresponds to a material whose thermal diffusivity is 20% and thermal conductivity is 24% of copper, while the volumetric heat capacity is approximately 120% of copper. This configuration lies within the physically admissible material cluster in **Fig. 8** and represents a realistic yet nontrivial identification target, since the sample differs substantially from copper while remaining relevant to high conductivity materials. The goal is to recover these properties using only temperature measurements taken in the copper substrate, with no direct sensing access to the sample region.

**Fig. 9** illustrates the performance of the two-stage optimization approach. In Stage A, Instance 2 is trained to fit the sensor data and establish an accurate interface temperature profile $\bar{u}(1^+, \bar{t})$. Subsequently, Stage B focuses on optimizing Instance 1 to identify the thermophysical parameters $(\bar{\alpha}, \bar{\kappa})$. All results reported in **Fig. 9** represent the mean behavior over six independent runs with different random initialization seeds. The parameter evolution exhibits distinct convergence characteristics. The thermal diffusivity $\bar{\alpha}$ (**Fig. 9d**) demonstrates rapid convergence, decreasing sharply from the initial estimate of 0.9 to approach the true value of 0.2 within the first 500 epochs, ultimately achieving a relative error below 5%. In contrast, the thermal conductivity $\bar{\kappa}$ exhibits a more gradual descent, converging to 0.24 with a final error of less than 1%. The derived volumetric heat capacity $\bar{\rho}\bar{c}$ (**Fig. 9f**) stabilizes around 1.25, confirming thermophysical consistency of the identified parameters. The robustness of this approach to different initial parameter guesses is further validated through ablation studies in **Figure S2** in the Supplementary Materials.

Statistical analysis shown in **Fig. 9g** reveals good accuracy and stability across all parameters. The thermal diffusivity $\bar{\alpha}$ achieves a mean error of 4.80% with a standard deviation of $\pm 2.83\%$, indicating consistent identification performance. The thermal conductivity $\bar{\kappa}$ demonstrates superior accuracy with only 0.71% mean error and good stability with a $\pm 1.93\%$ standard deviation. The volumetric heat capacity $\bar{\rho}\bar{c}$ maintains a 4.28% mean error with a $\pm 3.70\%$ standard deviation. These results confirm that all parameters are identified with high accuracy (mean errors below 5%) and good stability (standard deviations below 4%), with $\bar{\kappa}$ showing particularly robust performance. All recovered parameters therefore remain within 5% mean error and 4% standard deviation, with conductivity exhibiting particularly strong concentration, indicating that the HTF-based framework provides reliable and repeatable identification.

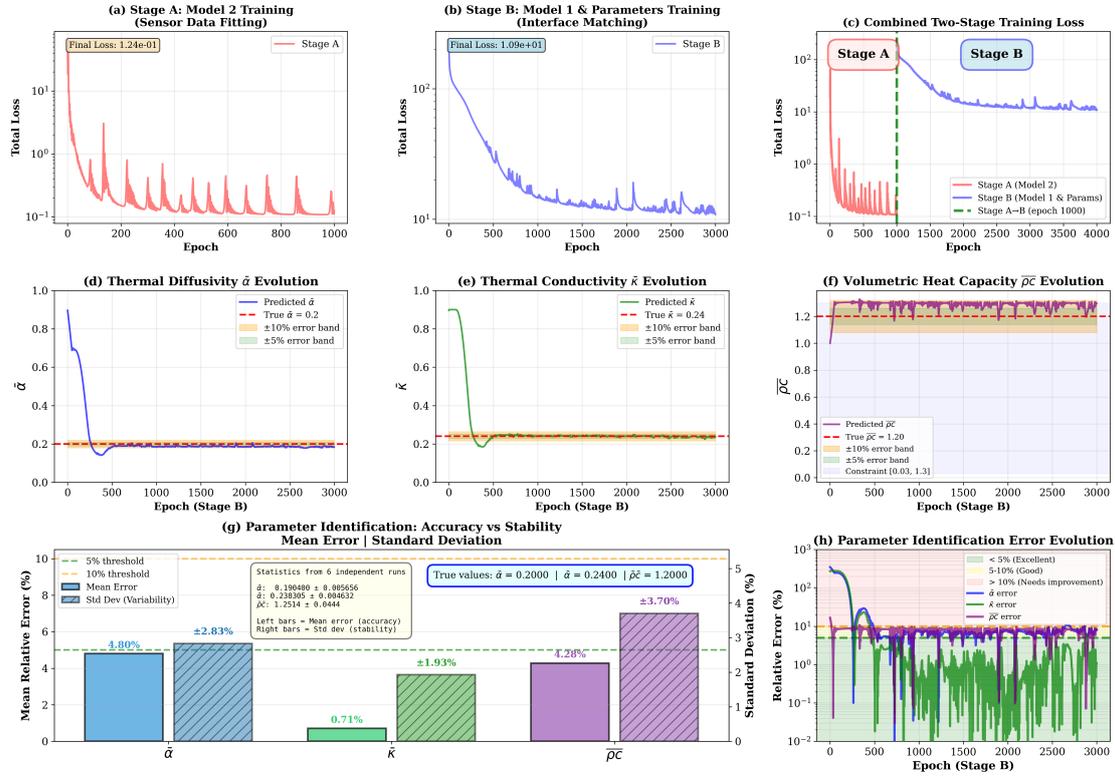

**Figure 9.** Performance analysis of the two-stage optimization strategy. (a-b) Training loss evolution for Stage A and Stage B. (c) Combined loss trajectory showing the transition at epoch 1000. (d-f) Parameter evolution curves for thermal diffusivity $\bar{\alpha}$, thermal conductivity $\bar{\kappa}$, and volumetric heat capacity $\bar{\rho}\bar{c}$, demonstrating convergence to true values ($\bar{\alpha}$ = 0.2, $\bar{\kappa}$ = 0.24, $\bar{\rho}\bar{c}$ = 1.2) with error bands indicating $\pm 5\%$ (yellow) and $\pm 10\%$ (orange) bounds. (g) Final identification accuracy: $\bar{\alpha}$ achieves 4.80% mean deviation, $\bar{\kappa}$ achieves 0.71%, and $\bar{\rho}\bar{c}$ achieves 4.36%, with standard deviations (hatched bars) indicating parameter stability. (h) Relative error evolution in Stage B showing all parameters stabilizing below 10% error, validating the effectiveness of the two-stage approach

in decoupling parameter identification despite measurement noise and the ill-posed nature of the inverse problem.

To understand why the method succeeds, it is essential to recognize a core difficulty of inverse heat conduction that is often overlooked. The dimensionless diffusion equation

$$\frac{\partial \bar{u}}{\partial \bar{t}} = \bar{\alpha}\frac{\partial^2 \bar{u}}{\partial \bar{x}^2}$$

suffers from an inherent degeneracy: any rescaling pair $(c\bar{u}, \frac{\bar{\alpha}}{c})$ in

$$\frac{\partial (c\bar{u})}{\partial \bar{t}} = \frac{\bar{\alpha}}{c}\frac{\partial^2 (c\bar{u})}{\partial \bar{x}^2}$$

satisfies the same PDE, creating infinite mathematically valid but physically distinct solutions. This scaling symmetry typically causes parameter drift and convergence to unrealistic values. Our two-stage approach breaks this symmetry through reference material calibration, leveraging the HTF algorithm's demonstrated capability for accurate heat flux gradient computation as validated in **Section 4**.

Stage A establishes the absolute thermal scale by training Instance 2 on sensor data from the substrate with known properties, fixing the interface temperature $\bar{u}(1^+, \bar{t})$ by measured values. Stage B then transmits this calibration to Instance 1 through interface physics. Temperature continuity

$$\bar{u}(1^-, \bar{t}) = \bar{u}(1^+, \bar{t})$$

prevents arbitrary rescaling of $\bar{u}$, while flux continuity

$$\bar{\kappa}\frac{\partial \bar{u}(1^-,\bar{t})}{\partial \bar{x}} = \frac{\partial \bar{u}(1^+,\bar{t})}{\partial \bar{x}}$$

anchors $\bar{\kappa}$ to the calibrated reference. These interface constraints, combined with the HTF's precise gradient calculations, transform the ill-posed inverse problem into a well-constrained boundary value problem, uniquely determining $\bar{\alpha}$ and $\bar{\kappa}$ relative to physical scales defined by the reference material.

## 6. Conclusion

This work tackles a challenge in interfacial heat conduction by providing a unified approach that resolves the coupled temperature dynamics across heterogeneous materials while enabling reliable recovery of unknown thermal properties. We introduce HeatTransFormer as a physics-guided Transformer architecture designed to capture the sharp spatial variations and cross-interface coupling that arise near material boundaries with high fidelity. Through an anchor-based spatiotemporal sampling strategy, the LaplaceAct activation function, and a mask-free self-attention mechanism that supports bidirectional spatiotemporal coupling, HeatTransFormer can produce temperature fields that remain accurate and physically consistent in the presence of sharp interface gradients and strong thermal contrast.

Beyond forward modeling, HeatTransFormer enables reliable inverse identification of thermal properties, a task that has remained difficult due to the intrinsic scaling symmetry of the heat diffusion equation. The two-stage optimization strategy used in this work establishes a physically calibrated thermal scale within the reference material and transfers this calibration to the sample region through interface continuity conditions. This approach transforms an inherently ill-posed inverse problem into a well-posed and physically grounded identification process and enables precise recovery of thermal diffusivity, conductivity and volumetric heat capacity in regions where direct measurement is not feasible.

Overall, by combining domain knowledge with attention-based modeling, HeatTransFormer demonstrates that neural networks can be tailored to capture the essential characteristics of interfacial and diffusion-driven physical systems with both accuracy and interpretability. The principles used in HeatTransFormer provide a practical pathway for extending physics-guided Transformer models to a broader class of diffusion-driven and interface-influenced problems.

**Declaration of competing Interests**

The author(s) declared no potential conflicts of interest with respect to the research, authorship, and/or publication of this article.

**Acknowledgement**

The authors acknowledge the financial support from the NSF of China (Grant No. 12421002). Dr. Sha sincerely appreciates the support provided by the Postdoctoral Fellowship Program of CPSF (Grant No. GZC20240975) and the Shanghai Postdoctoral Excellence Program (Grant No. 2024244). The machine learning training experiments were conducted on the computing platform at the International Center for Applied Mechanics in Energy Engineering (ICAMEE), Shanghai University.

**Data availability**

The Python scripts for model training, analytical solution implementations, and post-processing code are available from the corresponding author upon reasonable request.

*Supplementary Materials of Modeling and Inverse Identification of Interfacial Heat Conduction in Finite Layer and Semi-Infinite Substrate Systems via a Physics-Guided Neural Framework*

[1]Wenhao Sha, [1,2,3][1]Tienchong Chang

[1]Shanghai Institute of Applied Mathematics and Mechanics, Shanghai Key Laboratory of Mechanics in Energy Engineering, Shanghai Frontier Science Center of Mechanoinformatics, School of Mechanics and Engineering Science, Shanghai University, Shanghai 200072, China

[2]Shanghai Institute of Aircraft Mechanics and Control, Tongji University, Shanghai 200092, China

[3]Joint-Research Center for Computational Materials, Zhejiang Laboratory, Hangzhou 311100, China

---

[1] To whom correspondence should be addressed, E-mail: tchang@staff.shu.edu.cn

## 1. Visualization of Results under Different ICs and BCs

To assess the versatility and accuracy of the HTF framework in modeling temperature distributions in layer and substrate model, we evaluate three chip-level heat transfer scenarios [1] of increasing complexity with the same physical properties specified in **Section 3.1**, as shown in **Fig. S1**, where **(a-c)** depict HTF predictions and **(d-f)** present the corresponding analytical solutions for validation.

The first case considers a problem with different initial temperatures, where the temperatures of the finite and semi-infinite regions are given by $\bar{u}(\bar{x}, 0) = 1.0$ and $\bar{u}(\bar{x}, 0) = 0.0$, respectively. The left boundary is maintained at a constant temperature $\bar{u}(0, \bar{t}) = 1.0$. This setup mimics a pre-heated chip being brought into contact with a cooler substrate during startup, where mismatched thermal histories drive rapid initial heat flow. The results in **Fig. S1 (a, d)** reveal the rapid dissipation of the sharp initial temperature discontinuity, with HTF accurately capturing the steep transient gradients and their smooth relaxation towards equilibrium, while ensuring both temperature and heat flux continuity across the interface.

The second scenario examines the solution of the problem with a variable surface temperature, where the left boundary condition is defined as a linear function of time, $\bar{u}(0, \bar{t}) = \bar{t}$. This configuration represents chip operation under controlled heating ramps, as seen in burn-in testing or gradual thermal loading during sustained high-power operation. As shown in **Fig. S1 (b, e)**, HTF accurately captures the heat conduction process in which temporal variations in the linearly increasing surface temperature drive spatial redistribution of heat within the domain, effectively tracking the progressive penetration of thermal waves while ensuring that temperature profiles remain physically consistent across the interface throughout the heating period.

The third scenario addresses the solution of the problem with a heat source $w = $ const. The problem is complicated by the introduction of a positive heat source with the strength $w$, where the differential equation for the finite part is of the form: $\frac{\partial \bar{u}(\bar{x},\bar{t})}{\partial \bar{t}} = \bar{\alpha} \frac{\partial \bar{u}^2(\bar{x},\bar{t})}{\partial \bar{x}^2} + \frac{w}{c\rho}$. This configuration corresponds to chip regions experiencing persistent internal power dissipation, leading to localized hotspots in processors under continuous heavy workloads. The results in **Fig. S1 (c, f)** demonstrate HTF's capability to model the steady buildup of spatially non-uniform temperatures under constant volumetric heating, accurately balancing the competing effects of internal heat generation, conduction, and interface transfer to produce physically realistic steady-state and transient responses. Across all three scenarios, the predictive accuracy remains consistently high, which is further confirmed by the quantitative error distributions discussed below.

The error analysis underscores the robustness of the HTF approach. As shown in the pointwise L1 error surfaces in **Fig. S1 (g-i)**, most regions are deep purple, indicating minimal prediction errors, with pointwise values typically below 0.04 across the entire spatiotemporal domain. The few localized peaks of higher error, shown as bright yellow spots, appear primarily near boundaries and during the initial transient stage where thermal gradients change most rapidly. The pointwise L2 error maps in **Fig. S1 (j-l)** display similar spatial patterns, with squared errors consistently below 0.002 across most regions and no unexpected high-error zones. Despite the increasing physical complexity across the three scenarios, which progress from simple initial discontinuities to time-varying boundaries and internal heat generation, HTF maintains uniformly low prediction errors. These results confirm that HTF accurately captures multi-region heat transfer dynamics with high physical fidelity and computational efficiency, offering a reliable alternative to traditional numerical methods for diverse thermal scenarios.

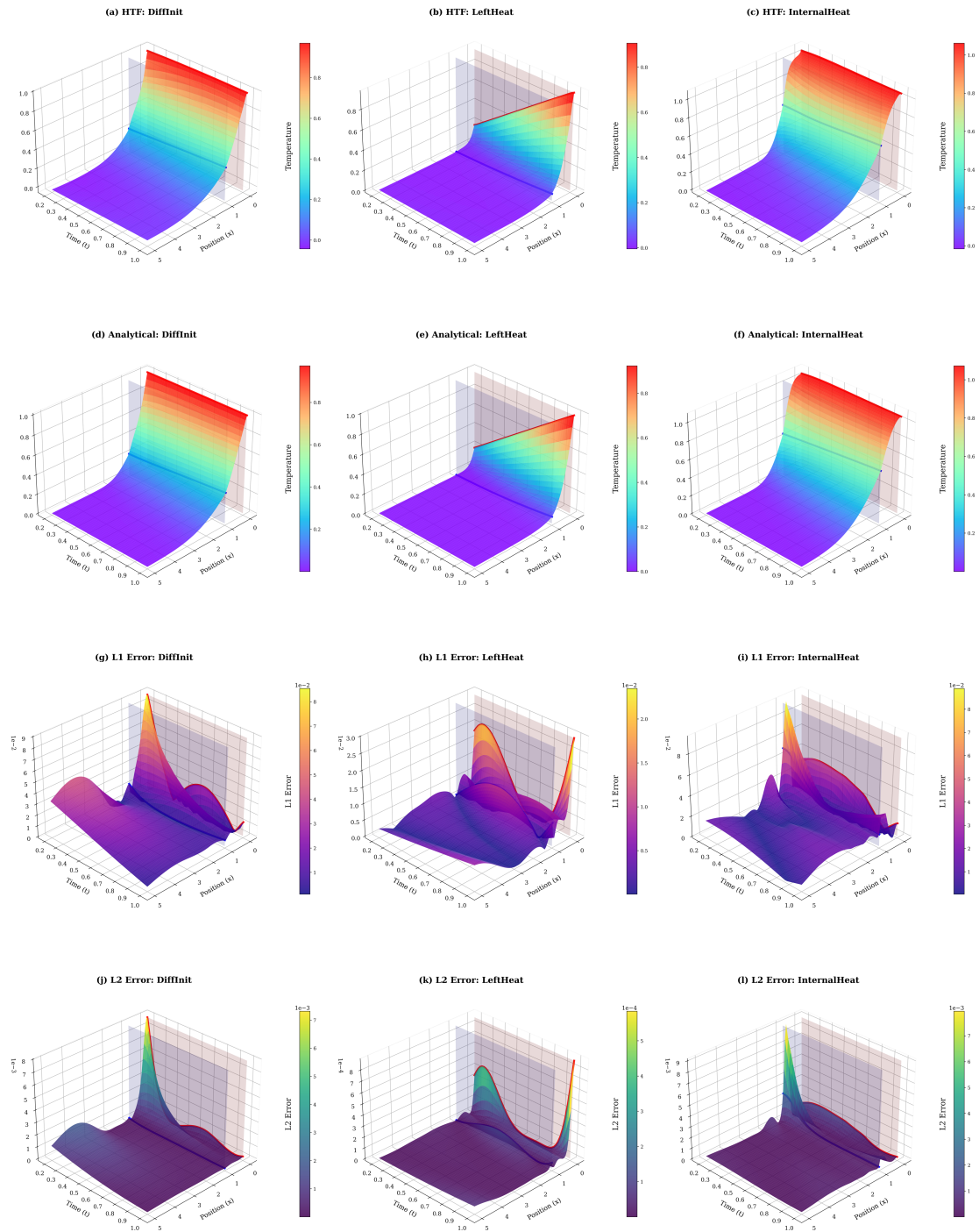

**Figure S1.** Comprehensive comparison of HTF predictions and analytical solutions for three heat transfer scenarios with error analysis. (a-c) Predicted temperature fields by HTF for: (a) different initial conditions, (b) time-dependent Dirichlet boundary on the left, and (c) internal heat generation. (d-f) Corresponding analytical solutions for each scenario. (g-

i) L1error maps. (j-l) L2 error maps. All plots include reference lines and surfaces at $\bar{x} = 0$ (red, left boundary) and $\bar{x} = 1$ (blue, interface) to facilitate spatial interpretation. The rainbow colormap is used for temperature distributions, while distinct colormaps (plasma for $L_1$, viridis for $L_2$) are employed for error visualization. Note: The analytical solutions for these three cases can be found in [1]. Note: The error calculation method in **Fig. S1** is consistent with that in **Fig. 4**.

## 2. Activation Functions and Their Mathematical Forms

| Activation | Formula | Notes |
|---|---|---|
| **ErfcAct** | $\text{ErfcAct}(x) = \text{amp} \cdot \text{erfc}(\text{scale} \cdot x + \text{bias})$ | Appears in diffusion solutions |
| **ReLUAct** | $\text{ReLUAct} = \max(0, \text{scale} \cdot x)$ | Classical piecewise-linear |
| **TanhAct** | $\text{TanhAct}(x) = \tanh(\text{scale} \cdot x)$ | Smooth bounded nonlinear |
| **SinAct** | $\text{SinAct}(x) = \sin(\omega_0 \cdot x)$ | Periodic, commonly used in PINNs |

**Table S1**. Activation functions used in **Section 4.2**.

## 3. Sensitivity of Inverse Identification to Initial Parameter Guesses with Extremely Small Initial Guesses $(\bar{\alpha} = 0.1, \bar{\kappa} = 0.01)$

To ensure consistency with the noise model used in the main text, we apply the same perturbation scheme to the synthesized interface temperature measurements. Specifically, the clean temperature field is corrupted by additive Gaussian noise:

$\bar{u}_{\text{measured}}(\bar{x}_i, \bar{t}_j) = \bar{u}_{\text{True}}(\bar{x}_i, \bar{t}_j) + \varepsilon$, where $\varepsilon \sim N(0, \sigma^2)$, with $\sigma = 1\%$.

This ensures that the sensitivity analysis is conducted under realistic sensor uncertainty conditions identical to those in the main experiments. To evaluate the robustness of the proposed two-stage inverse identification strategy under highly unfavorable initialization, we conduct an ablation test in which the thermal diffusivity and conductivity are initialized to values that are five to ten times smaller than their true counterparts. **Fig. S2** presents the resulting training dynamics for the two-stage optimization procedure.

Despite the severe mismatch between the initial guesses and the true material parameters, the sequential training strategy maintains stable convergence behavior throughout both stages. The model rapidly fits the sensor measurements in Stage A, yielding a smooth and accurate interface temperature profile. Notably, the quality of the reconstructed interface temperature remains unaffected by the poorly chosen initial material parameters, demonstrating the decoupling advantage of the two-stage design.

After switching to Stage B, both parameters progressively approach their true values: The diffusivity $\bar{\alpha}$ exhibits monotonic and stable improvement, rising from the extremely small initial value of 0.1 and gradually converging toward the ground truth of 0.2. The conductivity $\bar{\kappa}$, initialized at 0.01, also improves steadily toward the true value of 0.24, albeit with a slightly slower rate due to its weaker influence on the interface temperature compared with diffusivity. No divergence, oscillation, or parameter blow-up is observed, even under this unfavorable initialization. The derived volumetric heat capacity $\bar{\rho}\bar{c} = \bar{\kappa}/\bar{\alpha}$ stabilizes near the expected value during Stage B, further confirming the physical plausibility of the recovered parameters.

These results demonstrate that the two-stage inverse framework is highly robust to poor initialization, successfully recovering the correct thermophysical parameters even when starting from values that differ by more than an order of magnitude from the truth. This experiment thus complements the main-text analysis and reinforces the stability and reliability of the proposed inverse identification strategy.

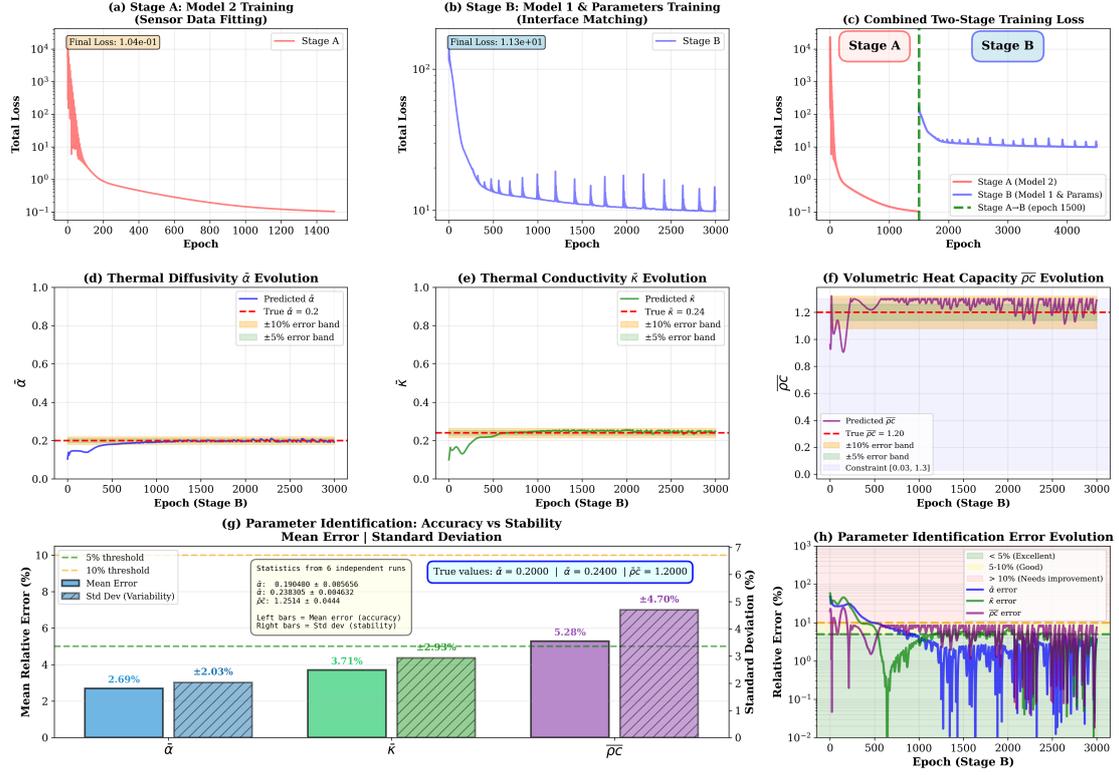

**Figure S2.** Performance analysis of the two-stage optimization strategy. (a-b) Training loss evolution for Stage A and Stage B. (c) Combined loss trajectory showing the transition at epoch 1500. (d-f) Parameter evolution curves for thermal diffusivity $\bar{\alpha}$, thermal conductivity $\bar{\kappa}$, and volumetric heat capacity $\bar{\rho}\bar{c}$, demonstrating convergence to true values ($\bar{\alpha}$ = 0.2, $\bar{\kappa}$ = 0.24, $\bar{\rho}\bar{c}$ = 1.2) with error bands indicating $\pm 5\%$ (yellow) and $\pm 10\%$ (orange) bounds. (g) Final identification accuracy: $\bar{\alpha}$ achieves 2.69% mean deviation, $\bar{\kappa}$ achieves 3.71%, and $\bar{\rho}\bar{c}$ achieves 5.28%, with standard deviations (hatched bars) indicating parameter stability. (h) Relative error evolution in Stage B showing all parameters stabilizing below 10% error, validating the effectiveness of the two-stage approach in decoupling parameter identification despite measurement noise and the ill-posed nature of the inverse problem.